\documentclass[twoside,leqno,singlecolumn,letter]{article}
\usepackage[colorlinks,linkcolor=cyan,citecolor=red,urlcolor=cyan]{hyperref} 
\usepackage{url,amsmath,amsfonts,mathtools,amssymb,graphics,graphicx,color, fullpage}
\usepackage[utf8]{inputenc}
\usepackage{authblk}
\usepackage{booktabs}
\usepackage{xr}
\usepackage{adjustbox}

\providecommand{\keywords}[1]{\textbf{\textit{Keywords---}} #1}

\usepackage{todonotes}
\usepackage{lscape}

\title{\bf{DeePathology: Deep Multi-Task Learning for Inferring Molecular Pathology from  Cancer Transcriptome}}

\author[1,4]{Behrooz Azarkhalili}
\author[2]{Ali Saberi}
\author[3]{Hamidreza Chitsaz}
\author[2, *]{Ali Sharifi-Zarchi}
\affil[1]{Department of Stem Cell Biology and Technology, Royan Institute, Tehran, Iran}
\affil[2]{Department of Computer Engineering, Sharif University of Technology, Tehran, Iran}
\affil[3]{Department of Computer Science, Colorado State University, Fort Collins, CO, USA}
\affil[4]{Department of Mathematics and Computer Science, Sharif University of Technology, Tehran, Iran}
\affil[*]{Corresponding author: asharifi@sharif.edu}
\date{}

\makeatletter
\def\@seccntformat#1{%
	\expandafter\ifx\csname c@#1\endcsname\c@section\else
	\csname the#1\endcsname\quad
	\fi}
\makeatother
\DeclareUnicodeCharacter{2223}{-}
\begin{document}
\maketitle 

\begin{abstract}\small
	
Despite great advances, molecular cancer pathology is often limited to the use of a small number of biomarkers rather than the whole transcriptome, partly due to computational challenges.
Here, we introduce a novel architecture of Deep Neural Networks (DNNs) that is capable of simultaneous inference of various properties of biological samples, through multi-task and transfer learning. It encodes the whole transcription profile into a strikingly low-dimensional latent vector of size 8, and then recovers mRNA and miRNA expression profiles, tissue and disease type from this vector. This latent space is significantly better than the original gene expression profiles for discriminating samples based on their tissue and disease.
We employed this architecture on mRNA transcription profiles of 10787 clinical samples from 34 classes (one healthy and 33 different types of cancer) from 27 tissues. 
Our method significantly outperforms prior works and classical machine learning approaches in predicting tissue-of-origin, normal or disease state and cancer type of each sample. For tissues with more than one type of cancer, it reaches 99.4\% accuracy in identifying the correct cancer subtype.
We also show this system is very robust against noise and missing values.
Collectively, our results highlight applications of artificial intelligence in molecular cancer pathology and oncological research.
DeePathology is freely available at \url{https://github.com/SharifBioinf/DeePathology}.
	
\end{abstract}

\section{Introduction}

Improving the accuracy of cancer diagnosis is extremely important for millions of patients and for far more non-patients who are tested worldwide every year. Despite great advances in oncologic pathology, there has been a significant ratio of errors that potentially affect the diagnostic results and/or treatment strategies \cite{Raab:2010gf}. Our systematic search through COREMINE (\url{www.coremine.com}) revealed 7652 article abstracts containing both   \emph{neoplasms} and \emph{diagnostic errors} (or synonymous terms), with an increasing trend over time. An M.D. Anderson Cancer Center study of 500 brain or spinal cord biopsies that were submitted to their neuropathology consultation service for a second opinion revealed 42.8\% disagreement between the original and the review diagnoses, including 8.8\% serious cases \cite{Bruner:1997wy}. A study of 340 breast cancer patients identified differences between the first and the second pathology opinions in 80\% of the cases, including major changes  that altered surgical therapy occurred in 7.8\% of cases \cite{Staradub:2002uw}. A review of 66 thyroid cancer patients revealed a different pathological diagnosis of 18\% of the cases \cite{Hamady:2005dx}. A recent study verified the accuracy and reproducibility of pathologists' diagnoses of melanocytic skin lesions for 240 skin biopsy cases from 10 US states and revealed 8-75\% error rates in different interpretation classes and an estimated 17.8\% whole-population error rate \cite{Elmore:2017fa}. Another recent study of 263 Australian Lichenoid keratosis patients revealed a diagnosis failure rate higher than 70\%, including 47\% of the cases misdiagnosed as basal cell carcinoma \cite{Maor:2017kt}. This situation is even worse in rare types of cancer. A study of 26 patients revealed 30.8\% misdiagnosis ratio in discriminating common gastric adenocarcinoma from hepatoid adenocarcinoma of the stomach, a rare subtype of gastric cancer \cite{Xie:2015vl}. Accurate diagnosis has been also challenging for a number of cancer types, including soft tissue sarcomas that are often misdiagnosed as other types of cancer \cite{Emori:2017ky}. 
\medskip

One limitation of the current molecular pathology methods such as Immunohistochemistry (IHC) is the limited number of genes or proteins monitored for diagnosis. Staining biopsies using antibodies against one or two proteins cannot discriminate between different cancer types if they have similar expression patterns of the target proteins. One possible solution is to use the whole-transcriptome of tissue biopsies \cite{Lapuk:2012gc}. But this approach is computationally challenging and different algorithmic and machine learning approaches have been employed so far to address this problem.  A subset of research is focused on \emph{binary classification}, e.g. discriminating between normal vs. tumor samples \cite{Li:2004cx, Bhat:2016us, MAHDIEH:2016wi, Yang:2016ha}. Stacked autoencoders are used for binary classification between glioma grades III vs. IV, and evaluated on 185 samples \cite{patil2018stacked}. These methods, however, can have limited clinical applications since most of the molecular pathology problems are \emph{multiclass}, e.g. assigning each sample to one of the different cancer types. Reaching high accuracy in classification problems usually becomes harder as the number of classes increases. Even a random assignment of samples to two classes will achieve 50\% accuracy if the classes are \emph{balanced} (i.e. there are an equal number of samples in each class), but a random classification will be around 3\% accurate if there are 33 balanced classes. Hence, it is important to consider the number of classes for comparing the accuracies of different techniques.
\medskip

Optimal Feature Weighting (OFW) is one of the earliest multiclass algorithms employed for cancer sample classification based on Microarray transcriptomes. This algorithm selects an optimal discriminative subset of genes and uses Support Vector Machines (SVM) or Classification And Regression Trees (CART). In previous work, it has been applied to five different problems, each consisting of 3 to 11 classes, without explicitly mentioning the obtained accuracies \cite{LeCao:2009gi}. A combination of SVM with Recursive Feature Elimination (RFE) is used to classify Microarray data of three cancer-related problems consisting of 3 to 8 classes, with accuracies between 95\% (for 8-class) to 100\% (for 3-class) \cite{Chai:ww}. Greedy search over top-scoring gene-sets has achieved an average 88\% accuracy, ranging from 48\% to 100\%, on seven different cancer datasets, each consisting of 3 or 4 classes with 40 to 96 samples per dataset \cite{Yang:2014fy}.

\medskip
One of the largest databases of cancer transcriptome, genome and epigenome profiles is Genomic Data Commons (GDC) that includes The Cancer Genome Atlas (TCGA) and Therapeutically Applicable Research to Generate Effective Treatments (TARGET) programs \cite{Grossman:2016ic}. There have been comprehensive works to analyze GDC data from different perspectives including identification of cancer driver somatic and pathogenic germline variations \cite{Ding:2018cz}, oncogenic signaling pathways \cite{SanchezVega:2018jw}, the role of cell-of-origin \cite{Hoadley:2018is} and cancer stem cells \cite{Malta:2018ic}, relationships between tumor genomes, epigenomes and microenvironments \cite{Ding:2018cz}. However, there has been little effort directed towards developing a molecular cancer pathology framework out of this valuable data. Semi-supervised stacked sparse autoencoders are used for binary classification of three types of cancer, and evaluated on 1311 samples \cite{xiao2018semi}. DeepGene, a feed-forward artificial neural network (ANN) based classifier, uses somatic point mutations profiles in order to assign each sample to one of 12 different cancer types. It achieved a mean 58\% and maximum 64\% accuracy over 3122 TCGA samples from 12 cancer types \cite{Yuan:2016eaa}. Another study used genetic algorithms and k-Nearest Neighbors (KNN) to classify TCGA samples based on RNA-seq transcriptome profiles. It reached about 90\% accuracy for classification of 602 normal samples and 9096 samples from 31 tumor types \cite{Li:2017dg}. 

\section{Methods}
\begin{figure}[h!]
	\begin{center}
		\includegraphics[width=.9\textwidth]{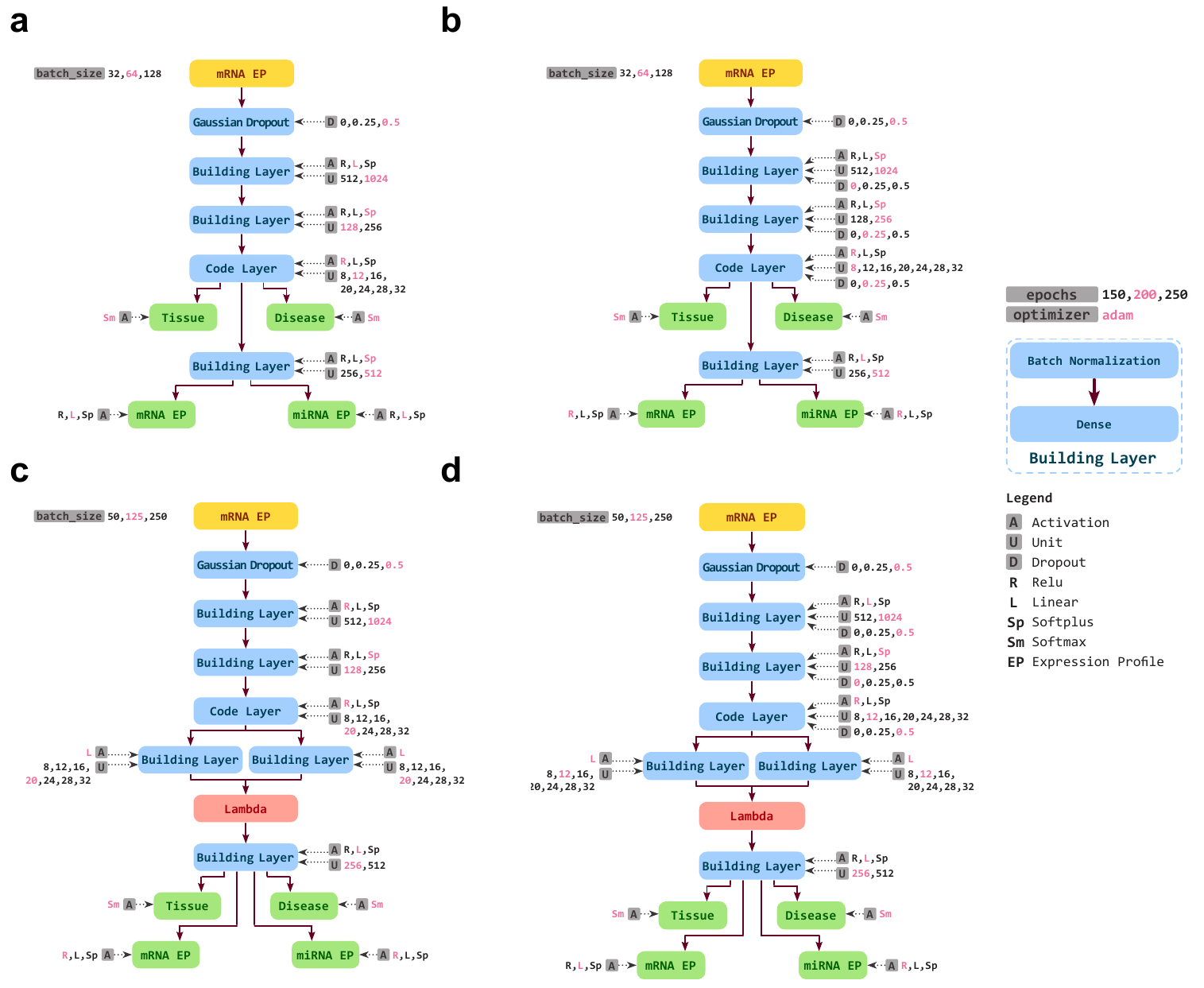}
	\end{center}
	\caption {Different architectures of DNN used for this study. \textbf{(a)} Contractive Autoencoder (CAE), \textbf{(b)} Dropout Contractive Autoencoder (Dropout-CAE), \textbf{(c)} Variational Autoencoder (VAE), \textbf{(d)} Dropout Variational Autoencoder (Droput-VAE). For each network, the layers are shown as boxes and the connections between them as arrows. The evaluated hyperparameters of each layer are shown next to each layer. The hyperparameter values in red show the optimal parameters identified through hyperparameter optimization. See extended methods for more details. }
	\label{fig:architectures} 
\end{figure}

\medskip
Here we used Deep Neural Networks (DNNs) in a multi-task learning approach to infer different biological and clinical information from transcriptome profiles. We designed four different architectures as shown in Fig.~\ref{fig:architectures} based on two different methods, including  Contractive Autoencoder (CAE) and Variational Autoencoder (VAE). 
Each of our DNNs consists of two parts that are serially connected:  the \emph{encoder} part, that learns to convert a given mRNA expression profile (mRNA EP) of a clinical sample at the input layer to a latent representation, which we call Cell Identity Code (CIC), and the \emph{decoder} part that infers multiple outputs from the CIC. While the CIC is a simple vector of numerical values in the CAEs, the VAEs encode the input into two equal-sized vectors, which represent means and standard deviations of multiple Gaussian distributions. For each method, we designed two architectures, one simpler and another consisting of dropout layer after each layer of the encoder. To make our architectures resistant to missing values and noisy data, we added a \emph{Gaussian Dropout} layer after the input, which perturbs the data with Gaussian noise and randomly sets some input values to zero. More details are provided in the online methods. 

\medskip
What makes our architectures different from conventional autoencoders is its particular design to learn \emph{useful} latent representations of the input data. Altered weights of the encoder part result in different latent representations of the same input, which can be decoded to the same output by different decoder weights. Hence, there is an infinitely large number of latent representation sets for a given set of data, however, the question is whether all of these representations are useful. Some latent representations might be extremely cryptic, while the others might be very useful for classification of biological samples, or obtaining other information. For instance, classification would be much easier if all cells of each type cluster together, distantly from the other clusters, in the latent representation space. The challenge is how to train the network in order to learn useful latent representations of the mRNA EPs. 

\medskip
To address this challenge, we designed the decoder part of each network to simultaneously learn four different classification and regression problems using the CIC, in a multi-task learning scheme: (i) reproducing mRNA  EP as one of the outputs that is as close to the original mRNA EP in the input as possible, (ii) predicting a miRNA expression profile (miRNA EP) that is as close as possible to the experimentally measured miRNA EP of the same sample, (iii) predicting the sample tissue of origin, among 27 different tissues, and (iv) predicting the sample disease state, which can be either normal or one of 33 different cancer types. All parts of each network, including encoder, decoder and classifiers were trained simultaneously.

\medskip
From the above tasks, (i) and (ii) can be viewed as non-linear regression, and (iii) and (iv) are classification. Importantly, the multi-task learning part of the DNN is aimed to accomplish all of these tasks only by getting the CIC as the input. Task (i) is to ensure the CIC stores much of the information in the original mRNA EP. Due to task (ii), we selected from GDC a subset of 10787 having both mRNA and miRNA EPs available. Furthermore, we removed 11 miRNA-encoding genes from the mRNA profiles, to make this task non-trivial. 

\medskip
The other key advantage of our model is \emph{hyperparameter optimization}. 
In addition to the internal network parameters (i.e. synapse weights and neuron bias values), each network has a set of hyperparameters, including the number of neurons in each layer, activation functions, size of mini-batches, standard deviations of the Gaussian noises, ratios of dropout layers, and the number of training epochs. Altered values of hyperparameters greatly affect the network results. Due to their cryptic inter-dependencies, all hyperparameters are required to be optimized simultaneously. For this purpose, we performed a comprehensive hyperparameter optimization through a Bayesian approach that runs in a number of iterations. In each iteration, it tries to find a set of hyperparameters that has the maximum likelihood of optimizing the network training outcome by integrating the results of all previous iterations in a Bayesian model. This process has superior advantages over grid search or random search of the hyperparameters in reducing the search space and more direct approaching the optimal hyperparameters using much fewer iterations.

\medskip
As shown in Fig.~\ref{fig:hyperopt_dis}, both network architecture and hyperparameters have a big impact on the regression and classification results. The red values in Fig.~\ref{fig:architectures} depict the optimal hyperparameter values. Surprisingly, a vector of length eight was optimal as the CIC of the Dropout-CAE, which outperformed the other networks in  regression and classification tasks. It means a set of eight numeric values is sufficient to represent almost the whole transcriptome profile and critical features of a biological sample.

\subsection{Data Pre-processing}

Transcriptome profiles of 11,500 samples were obtained from the Genomic Data Commons (GDC) consortium \cite{Grossman:2016ic}, including The Cancer Genome Atlas (TCGA) \cite{Ding:2018cz} and the Therapeutically Applicable Research to Generate Effective Treatments (TARGET) databases. For each mRNA profile, its corresponding miRNA profile was needed to train the model. So we kept the samples that both mRNA and miRNA expression profiles were available. This resulted in a total number of 10787 samples, including 10150 tumor and 637 normal samples. The mRNA and miRNA profiles of each sample were matched by the ID values of the patients, using the TCGAbiolinks R/Bioconductor package \cite{Colaprico:2016ep}. We randomly selected  about 10\% of the whole data as the test dataset, and the remaining samples were assigned to the training dataset.

One of our tasks was to predict expression level of miRNAs from mRNA profile. Hence we removed all miRNA genes from mRNA profiles, to ensure the answer to this task is not provided to the network as input. As the result, each mRNA and miRNA expression profile was of sizes 19671 and 2588, respectively. The precise number of samples from each type of tissue and tumor is provided in table \ref{tbl:numsamples}.

\subsection{Deep Autoencoder}
An interesting Neural Network (NN) architecture is autoencoder (AE), which compresses the input into a latent space representation, and then reconstructs the input back from this  representation. In other words, an AE seeks to learn an identity-like mapping function $f$ such that $f(x)\approx x$. AEs can reduce the dimensionality of data without losing significant information and can be trained using unlabeled data, hence they are widely used in different problems including data compression, dimensionality reduction, manifold learning, and feature learning \cite{Charte2018}.

\medskip
Each autoencoder consists of two parts, the \emph{encoder} and the \emph{decoder}, which can be defined as transition functions $f$ and $g$ such that:
\begin{eqnarray}
	f: \zeta \rightarrow \mathcal{F} \\ 
	g: \mathcal{F} \rightarrow \zeta \\
	f , g =\arg\min\limits_{\phi, \psi} \mathcal{J}(x, \tilde{x}(\phi, \psi, x))
\end{eqnarray}

where $\psi$ and $\phi$ are the encoder and the decoder functions, respectively, $\tilde{x}(\phi, \psi, x)=\phi\circ\psi(x)$ is the reconstruction of input vector $x$ and $\mathcal{J}=\sum\limits_{x \in D}L(x, \tilde{x}(\phi, \psi, x))$ is the total loss, which is evaluated as the summation of reconstruction error $L$ on training dataset $D$.

\subsection{Autoencoders variants}

While training an AE, we aim not only to reconstruct the input from the latent representation, but also to extract beneficial features in this representation. Therefore, it is vital to utilize some forms of regularization techniques to avoid overfitting or useless representations, even if the AE can reconstruct the input with minimal loss \cite{Charte2018, Rifai2011}. Different regularization techniques can veritably produce different variations of objective functions, and subsequently different features extracted from the data. The next sub-section explains the autoencoder variants of this study.

\subsubsection{Denoising Autoencoder}
A beneficial form of regularization is used in \emph{denoising autoencoders} (DAE), where the input vector $x$ is slightly corrupted and the autoencoder is expected to reconstruct the clean data from the latent representation \cite{Vincent2008}. As a result, the DAE learns to resist against input noise and overfitting. The following objective function is used for training DAEs:
\begin{eqnarray}
\mathcal{J}=\sum\limits_{x \in D}\mathbb{E}_{\hat{x}\sim q(\hat{x}|x)}L(x, \tilde{x}(\phi, \psi, \hat{x}))
\end{eqnarray}
where the expectation is evaluated over the corrupted versions $\hat{x}$ of the original data $x$, obtained from a corruption function $q(\hat{x}|x)$. This objective is optimized by stochastic gradient descent (SGD) or another iterative optimization algorithm. Additive isotropic Gaussian noise and binary masking noise are among the most frequently used corruption processes.

\medskip
Dropout \cite{Goodfellow2016, Srivastava2014} is another regularization technique that introduces some noise to the nodes  of any hidden layer, in contrast to the DAE that adds noise only to the input layer. The foremost dropout techniques utilized in deep learning are Bernoulli and Gaussian. In the former case, the output values of individual nodes are either dropped to zero with a probability $1-p$, or kept unchanged with a probability $p$. In the latter case, a multiplicative one-centered Gaussian noise $\mathcal{N}(1, \sigma^2)$ is applied to the output values of the nodes. 

\subsubsection{Contractive Autoencoder}

An alternative regularization technique is used in contractive autoencoder (CAE). If $h=\psi(x)$ is the latent representation of the input data $x$, the regularization term in CAE is the total squares of all partial derivatives of $h$ with respect to each dimension of the previous layer; therefore, the objective function of a CAE is expressed as:

\begin{eqnarray}
\mathcal{J}=\sum\limits_{x \in D}L(x, \tilde{x}(\phi, \psi, x)) +\lambda \|J_{\psi}(x)\|_F^2
\end{eqnarray}

Where the penalty term $\|J_{\psi}(x)\|_F^2$ is the Frobenius norm of the Jacobian matrix of the encoder activations with respect
to the input, and $\lambda$ is a balancing factor. The main goal of this term is to enforce the learned representation to be robust against small variations in the input data.

\subsubsection{Variational Autoencoder}
Variational autoencoder (VAE) is a special type of autoencoder with additional constraints on the encoded representations. It assumes that a latent, unobserved random variable $z$ exists, which
can leads to the observations $x$ by some stochastic mapping. As a result, 
its objective is to approximate the distribution of the
latent variable $z$ given the observations $x$.
\medskip

VAEs replace deterministic functions in the encoder and decoder by stochastic mappings; and compute the objective function in virtue of the density functions of the random variables:

\begin{eqnarray}
J( \phi, \theta,x)=D_{KL}(q_{\phi}(z∣x)\|p_{\theta}(z))-\mathbb{E}_{q_{\phi}(z∣x)}(\log(p_{\theta}(x|z)))
\end{eqnarray}

Where $D_{KL}$ stands for the Kullback–Leibler divergence, and $q$ is the distribution approximating the true latent distribution of $z$, and $\theta$, $\phi$ are the parameters of each distribution. The prior distribution over the latent variables is generally set to standard multivariate Gaussian $p_{\theta}(z) = \mathcal{N}(0,\sigma^2I)$; however, alternative distribution have also been recently considered.
\medskip

Although AEs can not generally be able to construct meaningful outputs from arbitrary encodings, VAE can learn a model of the data that can generate new samples from scratch by random sampling from the latent distribution. Therefore, VAEs are among the generative models.

\subsection{Models Architecture}
Our problem consists of two regression, namely reproducing mRNA and miRNA profiles from the latent variables, and two classification tasks to determine the tissue and disease state of each sample. All mRNA and miRNA expression profiles were normalized in $[0,1]$ by max-norm method of Scikit-Learn package  \cite{Pedregosa2012}. We subsequently constructed the multi-input and multi-output models in Keras \cite{Keras}. Keras is a high-level Python NN library that runs on top of either TensorFlow \cite{Abadi2016} or Theano \cite{TheTheanoDevelopmentTeam2016}.

\medskip
We constructed four different NN architectures: 

\begin{itemize}
	\item Variational autoencoder (VAE)
	\item Dropout-VAE, an extension of VAE with Bernoulli and Gaussian dropout layers being utilized for denoising.
	\item Contractive autoencoder (CAE)
	\item Dropout-CAE, the extended CAE with Bernoulli and Gaussian dropout layers.
\end{itemize}

To make each network tolerate input noise, an optional Gaussian noise $\mathcal{N}(0,\ \sigma^2)$ can be added to the input. 

\subsection{Loss Functions}

We utilized \emph{cosine similarity} as the loss for the classification tasks. Another choice was \emph{categorical cross entropy}, but we preferred to use the former function since it provides bounded results, which can be easily used in hyperparameter optimization.  We also used \emph{mean squared error (MSE)} as the loss function for the regression tasks. The total loss in our multi-task learning problem was be the weighted sum of individual losses, where the weights were $5\times 10^{-1}$ for each classification task, and $1\times 10^{-3}$ for each regression task. To evaluate the model performance in the validation dataset, we used used accuracy as the classification metric, and MSE and mean absolute error (MAE) for the regression tasks.

For presentation of the results, we preferred to use balanced accuracy. For imbalanced datasets, the accuracy can be misleading. For instance if 95\% of the data are normal, classifying all samples as normal gives 95\% accuracy. Balanced accuracy solves this problem by normalizing the number of correctly predicted samples of each class by the size of the same class, as below:

\begin{eqnarray}
\text{Balanced accuracy} = \frac{1}{2}(\frac{\text{True positives}}{\text{Condition positives}}+\frac{\text{True negatives}}{\text{Condition negatives}})
\end{eqnarray}

\subsection{Batch Normalization}
A major issue in training the DNNs is altered distribution of each layer inputs, as the weights and other parameters of the previous layers change. As a result of this \emph{internal covariate shift}, the learning rate should be lowered that causes reduced training speed rate, and also the initialization parameters should be assigned carefully \cite{Ioffe2015}. A well-known approach to address these challenges is \emph{batch normalization}. 

\medskip
Let's $x_{1\cdots m}$ be the values of an activation $x$ during a mini-batch. Then the batch normalized values $y_{1\cdots m}$ are computed as follows:

\begin{eqnarray}
y_i = \gamma \frac{x_i-\mu}{\sqrt{\sigma^2+\epsilon}}+\beta
\end{eqnarray}

where $\mu$ and $\sigma^2$ are the mean and variance of the $x_{1\cdots m}$ values, $\gamma$ and $\beta$ are the scaling parameters to be learned, and $\epsilon$ is a small constant value added to the mini-batch variance for numerical stability.

\medskip
This method basically standardizes the inputs of each layer in such a way that they have a mean and standard deviation of zero and one, respectively. Then scales the standardized values by a linear function, with parameters that are learned. Batch normalization is analogous to how the inputs to the networks are standardized, but it can be performed for each internal layer of a DNN. It turns out that, extending this technique to hidden layers can significantly improve the training speed. 

\medskip
As depicted in Fig.~\ref{fig:architectures}, each \emph{building layer} of our network is consisted of a batch normalization layer, followed by a dense layer. Our experience showed the batch normalization layers significantly improve the training speed of the networks.

\subsection{Hyperparameters Tuning}
Many machine learning methods have a set of architectural parameters, called \emph{hyperparameters}, which are determined prior to training the model. For example, the number of layers, the number of neurons per layer, the type of activation functions, and the type of dropout or noises are among the DNN hyperparameters. Since the values of hyperparameters affect the  model architecture, hyperparameter optimization can be considered as model selection technique. 

\medskip
A key advantage of our work is optimizing the network hyperparameters in order to achieve the best classification and regression objectives. Besides two simple methods of hyperparameter tuning including grid search and random search, there exists a more advanced Bayesian algorithm \cite{Shahriari2016}.

\medskip
Bayesian optimization, in contrast to random or grid search, keeps track of the past evaluations and utilizes them to define a probabilistic model mapping hyperparameters to a probability distribution of the outcome of the objective function $P(s | h)$, where $s$ and $h$ are the objective function score and hyperparameters. This is called a \emph{surrogate} model for the objective function, and is easier to be optimized in comparison with the objective function itself. In each iteration, the Bayesian optimization selects optimal hyperparameters based on the surrogate function as the next set of hyperparameters to be evaluated by the actual objective function. Briefly, it works as the following algorithm:

\begin{enumerate}
	\item Build a surrogate probability model of the objective function.
	\item Find the hyperparameters that perform best on the surrogate.
	\item Evaluate these hyperparameters by the actual objective function.
	\item Update the surrogate model by incorporating the new results.
	\item Repeat steps 2–4 for the specified number of iterations or running time.
\end{enumerate}

The main advantage of the Bayesian optimization is becoming more intelligent by continuously updating the surrogate probability model after each evaluation of the objective function. By intelligently selecting hyperparameters that are more likely to optimize the objective function in each iteration, they can find better set of hyperparameters in a fewer iterations, in comparison with random and grid search \cite{Dewancker2015, Snoek2012}. 

\medskip
To exploit Bayesian optimization in our model selection procedure, we utilized a Python module called Hyperopt \cite{Bergstra2013}. Hyperopt provides efficacious algorithms and parallel infrastructure. We needed to define a search space and an objective function for the hyperparameter optimization. The next sub-section explains the search space. As the objective function, we used an average of the total losses of all data, which was split to 80\% training and 20\% test dataset.

\medskip
While we benefited from all data in hyperparameter optimization, we performed 5-fold cross-validation while measuring classification accuracy to ensure the hyperparameters are not overfitted to some specific portion of the data.

\subsection{Hyperparameters Search Space}
For each architecture, we selected a set of hyperparameters; and for each hyperparameter, we defined a set of discrete values as the search space. The Cartesian product of all of these search spaces was used as the total search space of the hyperparameter optimization process. 

Below is a description of the hyperparameters:
\begin{itemize}
	\item{\textbf{Units}:}
	A critical feature of each layer is the number of its neurons. Based on our experiences in a prior work, we selected a wide range of different values as the search space of each layer. These values were selected in a decreasing order for the encoder part, to gradually narrow it down from the input to the latent (code) layer, followed by an increasing order in the decoder part.
	\item{\textbf{Activation Functions}:} 
	A widely used activation function is Rectified Linear Unit (ReLU), which is defined as $f(x) = \max(x, 0)$ \cite{Nair2010}. Since gradients can readily flow whenever the input to the ReLU function is positive, gradient descent optimization is much simpler for ReLU than sigmoid activation functions. Although a few efficacious activation functions such as Swish \cite{Ramachandran2017a} and Selu \cite{Klambauer2017} have been recently introduced; However, there has been no significant and universal enhancement to utilize them instead of ReLu function.
	
	Despite promising features, ReLU can cause some difficulties in autoencoders; including gradient explosion and saturation. Hence we additionally used linear activation functions $f(x)=x$ and SoftPlus activation function $f(x) = \ln(1+\exp(x))$, as comprehensively discussed \cite{Glorot2011}. 
	
	\item{\textbf{Dropout Rate}:}
	Dropout layer plays an important role by acting as regularizer to prevent overfitting. It has one main tunable parameter, which is the noise rate. We chose $\{0, 0.25, 0.5\}$ as the set of noise rates. By putting 0 in the options, we provided Dropout-CAE and Dropout-VAE models. By this way, these models become more flexible to either exploit or not to exploit dropout layers into their own structures to achieve the maximum performance.
\end{itemize}
For each NN architecture, we set Hyperopt to iterate over 200 different networks and select the best one among them.

\subsection{Dimension Reduction Analysis and Comparison With Other Algorithms}
We observed the performance of our architecture in reducing the initial dimension of the data by visualizing Principal Components Analysis (PCA) and t-distributed Stochastic Neighbor Embedding (t-SNE) of the mRNA profiles as well as CICs (Fig.~\ref{fig:tsne_disease}). This was done by performing 5-fold cross-validation and using the CIC of each sample when it was used in the test-dataset. 

We compared the classification accuracy of the proposed deep learning method against other classification methods including \textit{KNN}, \textit{Extra Tree Classifier}, \textit{Random Forest Classifier},  \textit{Stochastic Gradient Descent (SGD) Classifier} and \textit{Support Vector Machine (SVM)}. We selected these algorithms because they were available for parallel execution in Scikit-Learn package \cite{Pedregosa2012}. Moreover, they are among the most efficient and versatile classification algorithms. The inputs of all models were the same mRNA expression profiles. Each other machine learning algorithm was trained to solve one of the classification problems (i.e. tissue or disease). 

\medskip
To ensure the hyperparameters of the other classifiers are selected properly, we performed $100$ iterations of hyperparameter optimization on each classification algorithm (Fig.~\ref{fig:cls_hyperopt}). The results are shown in Fig.~\ref{fig:classifiers}a,c. As shown, our method has outperformed the other classification algorithms in both tissue and disease prediction.

\subsection{Availability of Data and Source Codes}	
All data can be obtained from Genomic Data Commons (https://gdc.cancer.gov). 
All data pre-processing and analysis source codes are available at \url{https://github.com/SharifBioinf/DeePathology}.

\section{Results and Discussion}

\subsection{Performance of the four different DNN architectures on different tasks}
\medskip
We trained the hyperparameter optimized networks for 200 epochs using the training dataset. Fig.~\ref{fig:hyperopt_train} shows the performance of these networks on the training dataset during the training, and Fig.~\ref{fig:hyperopt_test} shows their performance on the test dataset. The CAE and Dropout-CAE architectures depicted the best performance in all four tasks. For the regression tasks (i.e. reproducing the mRNA and predicting the miRNA EP) the CAE architecture had a slightly better performance over the Dropout-CAE, but for the classification tasks, both networks performed very similarly.
Furthermore, there was no sign of overfitting for these architectures as the test dataset errors did not increase during the training. Also, all accuracies and MSEs reached stable levels, that showed 200 training epochs was sufficient.
\medskip
A striking observation was the size of CIC, determined by hyperparameter optimization. Although there were higher values up to 32 available as options for the size of code layer, all architectures were optimized by using smaller sizes between 8 to 20 (\ref{fig:architectures}). Particularly, the Dropout-CAE that outperformed other architectures was optimized by a CIC of size 8 (\ref{fig:architectures}b). 
This means that the original mRNA profile of size 19671 is encoded into an 8-dimensional latent space, and then all of the mRNA and miRNA expression profiles, tissue and disease state of the samples are obtained from this CIC space. 

\subsection{Samples are better discriminated by CICs rather than the original mRNA expression profiles}
\medskip
The TCGA data is integrated from different studies, hence one might ask whether the original mRNA expression profiles of each study are discriminated from the other studies due to batch effects. 
In that case, classification of different cancer types might only learn artificial patterns of batch effects that discriminate studies, rather than the real biological features of transcriptomes that discriminate tissues and cancer types. 

To address this question, we compared the original 19671-dimensional mRNA expression profiles against the 8-dimensional CIC feature space that we learn using Dropout-CAE. It is noteworthy that each sample in the dataset will be used exactly once as a test sample in our cross-validation procedure; hence, all the datasets utilized for visualization and statistical analysis have been resulted in combining all corresponding samples generated by each fold of cross-validation. For visualization of the samples in each space, we used Principal Components Analysis (PCA) as a linear dimension reduction algorithm, and t-distributed Stochastic Neighbor Embedding (t-SNE)  \cite{Maaten:2008tm} as a non-linear manifold-learning algorithm.

\begin{figure}[h!]
	\begin{center}
		\includegraphics[width=0.7\textwidth]{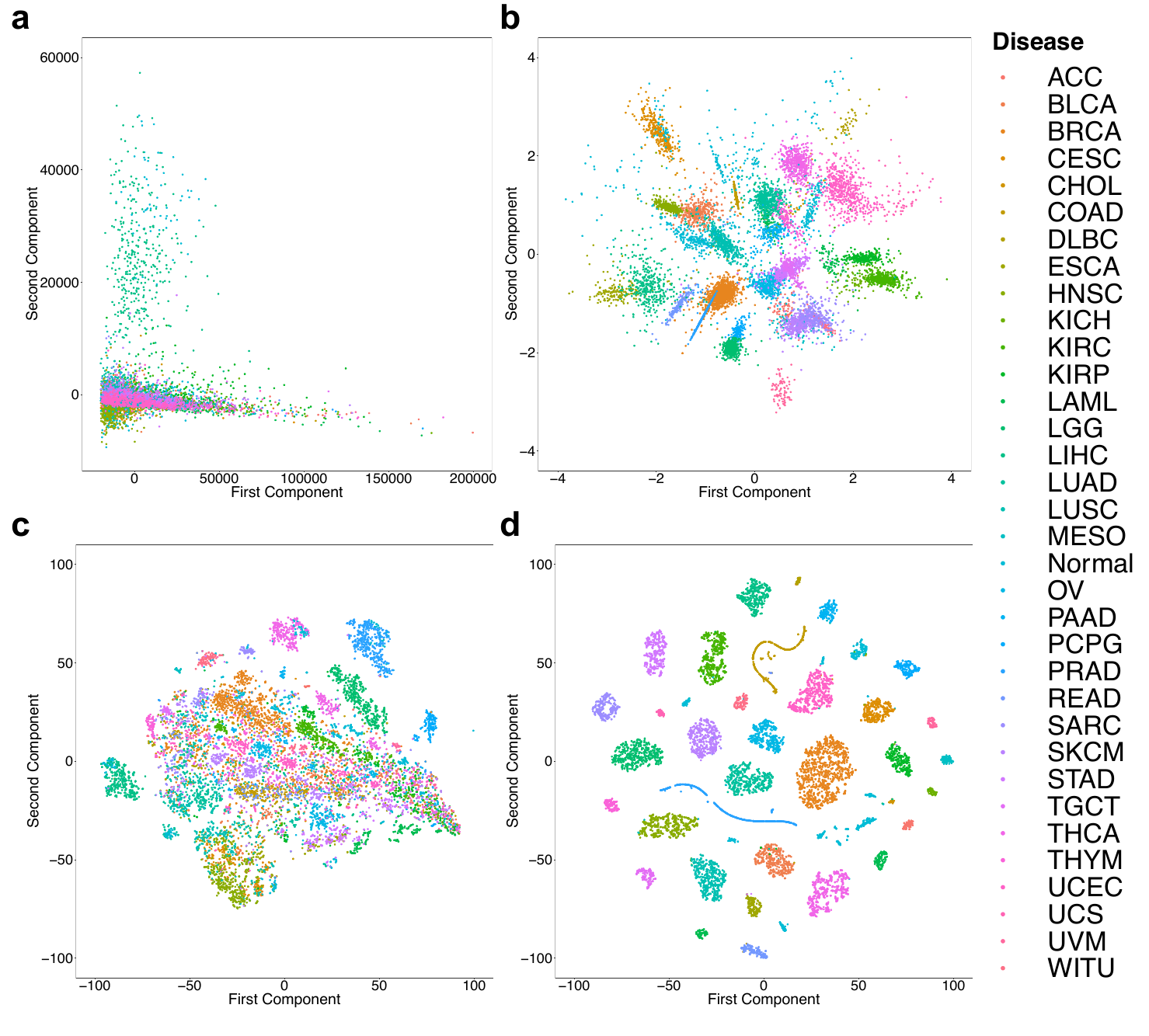}
	\end{center}
	\caption {Discrimination of disease state of samples using the original mRNA expression profiles vs. Cell Identity Codes (CIC). \textbf{(a)} PCA plot of the original mRNA expression profiles of all samples. Each dot and its color show a sample and its disease state, respectively. \textbf{(b)} PCA plot of the 8-dimensional CIC space.  \textbf{(c)} t-SNE plot of the original mRNA expression profiles.  \textbf{(d)} t-SNE plot of the 8-dimensional CIC space.
	}
	\label{fig:tsne_disease} 
\end{figure}

As shown in Fig.~\ref{fig:tsne_disease}a, the original mRNA expression profiles from most of the studies overlap and it's impossible to discriminate disease states of the samples from 2D PCA.  The same PCA projection, however, has a significantly better discrimination of disease states if applied to the 8-dimensional latent space of the CICs (Fig.~\ref{fig:tsne_disease}b). 
Furthermore, we used the 8-dimensional feature space obtained by different types of PCA to predict the tissue and disease state of each sample (Fig.~\ref{fig:pca}). For this purpose, we used an Ensemble learning method based on SVM and RandomForest that predict tissue or disease state based on the first 8 principal components of each sample. The maximum classification accuracies for tissue and disease were 38.7\% and 33.2\%, respectively. As we will see in the next section, these results were far below the ac curacies obtained by DNN using CICs.  
\medskip
A more recent manifold learning algorithm called t-SNE is known to provide better representation of biological samples such as transcriptomes or single-cell expression profiles in low-dimensional spaces than linear methods such as PCA. We compared t-SNE visualization of original mRNA expression profiles (Fig.~\ref{fig:tsne_disease}c) against latent CIC representations (Fig.~\ref{fig:tsne_disease}d). Evidently, the 8-dimensional CIC space is significantly better discriminative of the disease state of the samples than the original gene expression profiles. Doing the same analyses for studying tissue-type discrimination provided similar results (Fig.~\ref{fig:tsne_tissue}). 
\medskip
Taken together, these results have two important conclusions: (I) There is no significant batch effect in the original mRNA expression profiles that can discriminate different studies. (II) The 8-dimensional CIC space is learning features from the transcription profiles that can very well discriminate different tissues and diseases. 

\subsection{DNN outperforms other classification algorithms in identifying tissue type and disease state}
\begin{figure}[h!]
	\begin{center}
		\includegraphics[width=0.7\textwidth]{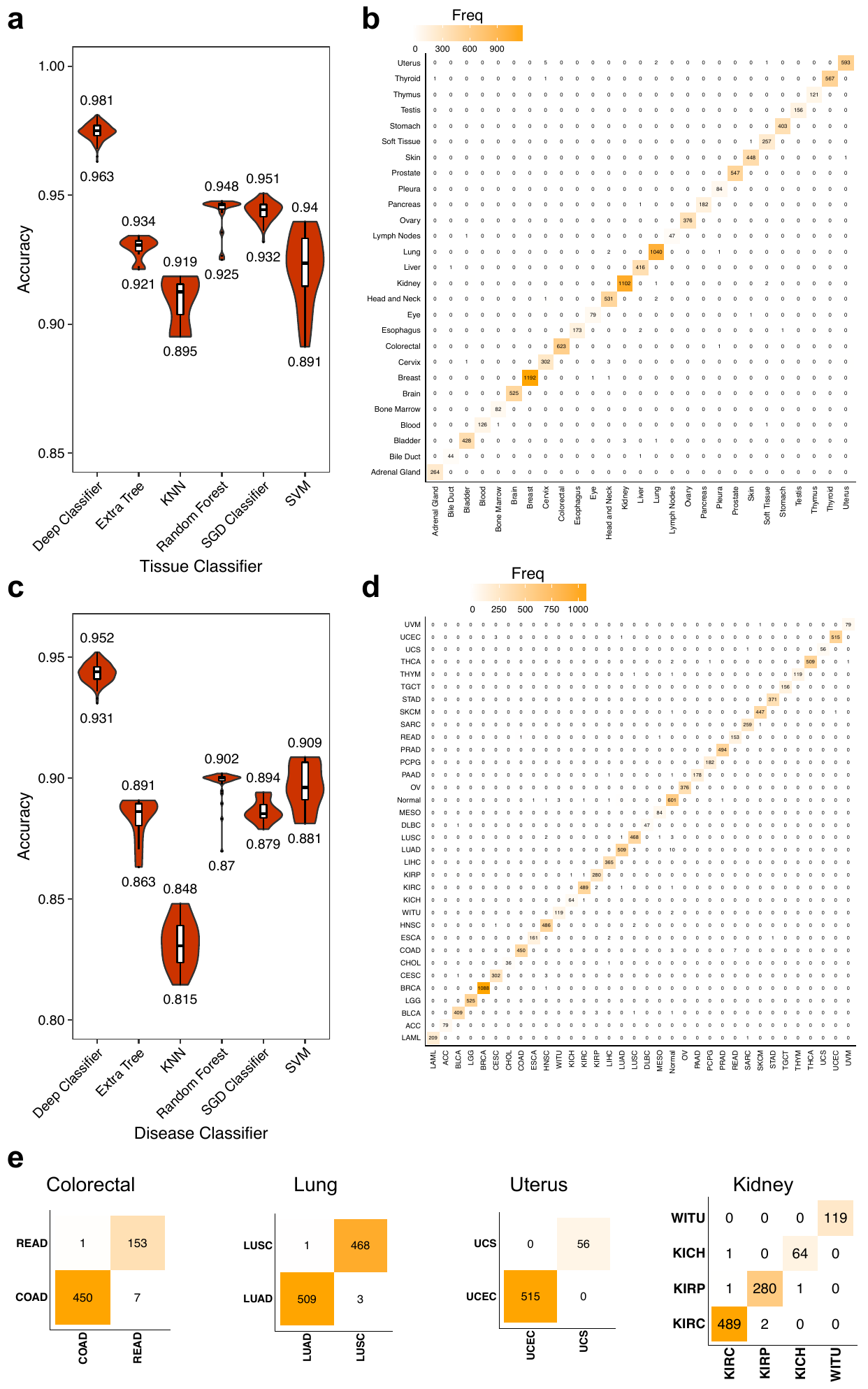}
	\end{center}
	\caption {Accuracy of the DNN, in comparison with other classifier algorithms. \textbf{(a)} Each violin shows the distribution of tissue classification accuracies obtained by hyperparameter optimization of different algorithms. The leftmost violin shows our algorithm, and the other violins show some widely-used classification algorithms. \textbf{(b)} Confusion matrix of tissue classification, in which the $x$ and $y$ axes represent the actual and predicted tissue types. Each cell contains zero or some positive number of samples, and the diagonal and other cells represent correct and incorrect classifications, respectively. \textbf{(c, d)}  Similar analyses for classification of the disease type.
	}
	\label{fig:classifiers} 
\end{figure}

%\section{Results}
\medskip
We compared the classification accuracy of our method with several other widely-used classification algorithms. For this purpose, we selected k-nearest neighbors (KNN), Extra Tree, Random Forest,  Stochastic Gradient Descent (SGD), and Support Vector Machine (SVM). To maintain a fair comparison, we also performed hyperparameter optimization for each of these algorithms. As shown in Fig.~\ref{fig:classifiers}a, our Dropout-CAE network outperformed the other classification algorithms in identifying the tissue of each sample. Fig.~\ref{fig:classifiers}b shows a tiny number of misclassifications in the confusion matrix of the Dropout-CAE. Figs.~\ref{fig:classifiers}c,d show similar results for cancer type classification.  Extended results are provided in Supplementing Tables~\ref{tbl:tissuef1} and \ref{tbl:diseasef1}.  It is also crucial to restate that all the datasets utilized for statistical analysis of  Tables~\ref{tbl:tissuef1} and \ref{tbl:diseasef1} have been resulted in combining all corresponding samples generated by each fold of cross-validation. 

\subsection{Stable classification of different tissues and cancer types}

\begin{figure}[h!]
	\begin{center}
		\includegraphics[width=\textwidth]{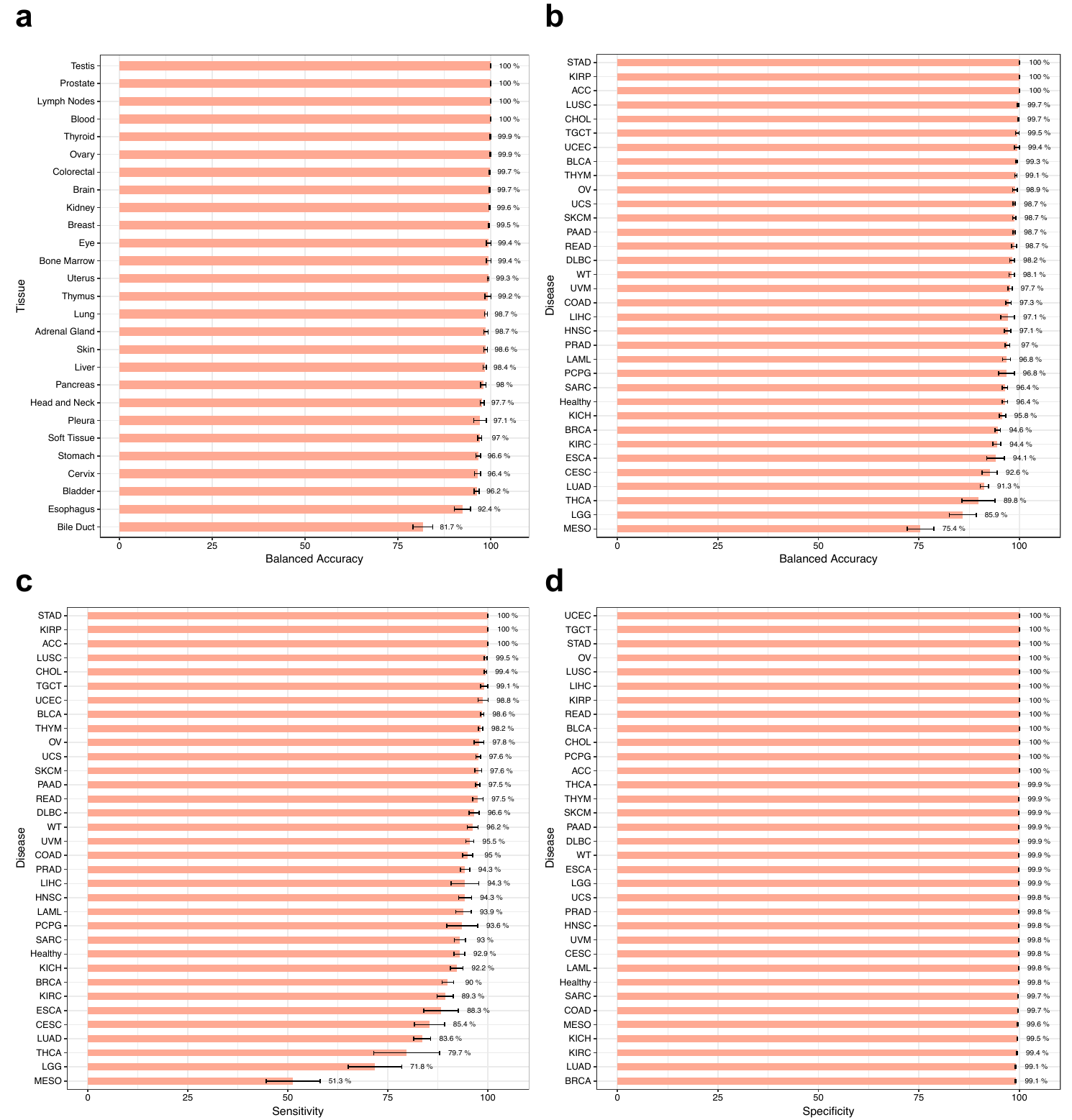}
	\end{center}
	\caption {Performance of DNN in the classification of different tissues and diseases. \textbf{(a)} Balanced accuracy of tissue classification.  \textbf{(b)} Balanced accuracy of disease classification. Healthy samples are depicted as "Healthy". \textbf{(c,d)} Sensitivity and specificity of disease classification. All analyses are performed using 5-fold cross-validation. 
		Error bars indicate standard error (SE).
	}
	\label{fig:disease_accuracy} 
\end{figure}

\medskip
To ensure the results are not overfitted to a particular subset of the data, we performed a 5-fold cross-validation and measured different accuracy criteria for tissue and disease classification. The data was shuffled and each sample was randomly assigned to one of 5 groups. In $i$-th round of cross-validation ($1\leq i \leq 5$), group $i$ was used as the test set and the remaining four groups were used for training. The results are depicted in Fig.~\ref{fig:disease_accuracy}.

\medskip
The balanced accuracy of tissue classification was $\geq$ 99\% for 14 tissues, and $\geq$ 95\% for 25 out of 27 tissues. This showed the DNN results are stable across different tissue types. Disease classification was $\geq$ 99\% accurate for 9 cancer types, and $\geq$ 95\% for 25 cancer types and normal tissues. Only 3 out of 33 cancer types had a balanced accuracy less than 90\%. The standard error among 5 rounds of cross-validation was negligible for most of the tissues and disease types. These analyses confirmed our method is not overfitted towards a particular tissue or disease type or some subset of the data. 

\medskip
Fig.~\ref{fig:disease_accuracy}c,d show sensitivity and specificity of DNN in disease classification. We observed an average sensitivity $\geq$ 99\% and $\geq$ 95\% for 6 and 18 cancer types, respectively. Seven cancer types had a sensitivity lower than 90\%. The specificity was $\geq$ 99\% for all 34 classes, including 33 cancer types and 1 normal tissues.

\medskip
It is important to mention that the samples considered as "Normal" in TCGA are not purely normal because most of them are obtained from the tissues adjacent  to cancer tumors. As a result, we can consider some samples that are labeled "Normal" to be predicted as non-normal in our model. This can be seen in Fig.~\ref{fig:classifiers}d, as the highest misclassfications have occurred in predicting "Normal" samples to be from one cancer type. It's hard to say that our model has made errors in these cases, and we expect the true accuracy of our model is slightly higher than the reported value. 

\subsection{DNN accurately discriminates different cancer subtypes}
\medskip
Different cancer types or subtypes might arise from the same tissue (e.g. lung squamous cell carcinoma vs. lung adenocarcinoma). Each cancer type/subtype might have a specific therapeutic strategy. Hence, correct discrimination of cancer types/subtypes that arise from the same tissue is a critically important and sometimes challenging task of cancer pathology.
\medskip
We focused on all 4 tissue types that had more than one type of cancer in this dataset: Colorectal, Lung, Uterus and Kidney. As shown in Fig.~\ref{fig:classifiers}e, the confusion matrices for these cancer types confirm there are only 17 misclassifications out of 3120 total samples, which provides us with 99.4\% accuracy for cancer  subtype classification. This accuracy ranges from 98.7\% (Colorectal cancer subtypes) to 100\% (Uterus). Discriminating cancer subtypes is clinically very important for deciding a correct therapeutic strategy, and is often challenging due to histological similarity of cancer subtypes and their heterogeneity. This finding paves the road of using high-throughput molecular data to address challenging pathology problems.

\subsection{DNN can resist noise and missing values}
\begin{figure}[h!]
	\begin{center}
		\includegraphics[width=0.7 \textwidth]{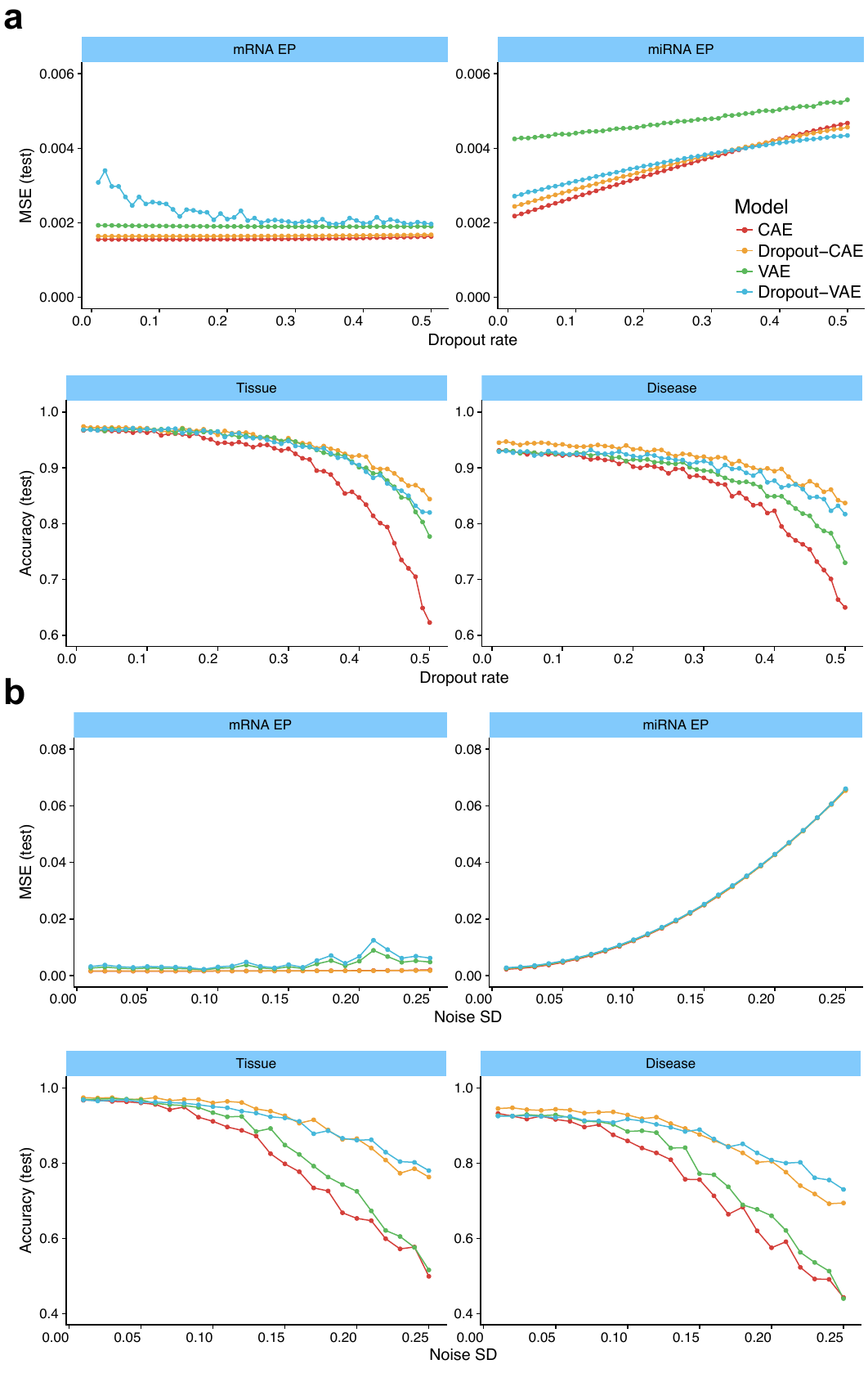}
	\end{center}
	\caption {The resistance of different DNN architectures against missing values and noise. \textbf{(a)} In each plot, the x-axis shows the fragment of randomly-selected input values that are set to zero (dropout), and y-axis shows either regression MSE or classification accuracy for the test dataset. \textbf{(b)} The x-axis shows the standard error (SD) of a zero-centered Gaussian noise which was added to the input values, and y-axis shows MSE or accuracy for the test dataset. Colors indicate different DNN architectures (see the figure legend). 
	}
	\label{fig:noise} 
\end{figure}

\medskip
A potential issue with classifiers is their reduced accuracy when the input data are noisy or have missing values, due to sample quality or measurement errors. To check the effect of missing values, we added a dropout layer after the input that randomly dropped the expression values of a random set of genes by setting them to zero. The fraction of dropout genes was increased from 0 to 50\% with 1\% steps. The results are presented in Fig.~\ref{fig:noise}a. In each plot, the $x$-axis shows the fraction of dropout genes, and the $y$-axis shows either regression MSE or classification accuracy for the test dataset. As shown, dropping the values of 50\% of the genes has a negligible effect on reproducing mRNA expression profiles, with almost no MSE change for all architectures excepting Dropout-VAE. Interestingly, the increased dropout rate caused an improved MSE for Dropout-VAE. Also, increasing the dropout rate elevated the MSE of predicting miRNA EP for all DNN architectures. But even at 50\% dropout, the MSE of all architectures was around 0.004, which is quite small. 

\medskip
The classification accuracies of all DNNs, particularly the Dropout-CAE and Dropout-VAE had small changes by increasing the missing values from 0 to 20\%. Even at 30\% dropout rate, Dropout-CAE was 95\% accurate. Disease classification accuracy of both Dropout-CAE and Dropout-VAE had also small changes by a dropout rate up to 20\%. These experiments showed that the DNN architectures that contained Dropout layer had the highest resistance to missing values.
 
\medskip
We also measured the resistance of DNNs to a noisy input (Fig.~\ref{fig:noise}b). A layer just after the input added  a zero-mean Gaussian noise to the input GEPs. The magnitude of the noise was controlled by increasing its standard deviation (SD) from 0 to 0.25 with 0.01 steps. Both CAE and Dropout-CAE architectures were quite resistant to noise in reproducing the mRNA EP, and their MSEs were almost unchanged at SD=0.25. All networks could predict miRNA EP with MSE $\le$ 0.01 when the SD of Gaussian noise was at most 0.08. But their errors were increased by increasing the noise magnitude. The tissue and disease classification accuracy of Dropout-CAE were almost unchanged for a noise with SD $\leq 0.05$. Dropout-CAE and Dropout-VAE outperformed the other architectures in resisting against Gaussian noise. 
 
 \medskip
Collectively, our results indicate the power of DNNs in obtaining biologically and clinically important information from transcriptome profiles. Our networks compress the transcriptome profile into a thumbnail CIC, and obtain tissue and disease type and miRNA expression profiles out of it. This process is greatly robust against noisy and missing data and outperforms baseline algorithms in accuracy. We suggest employing DNNs in inferring the outcome of molecular cancer pathology.

\section{Future works}

Due to the black-box nature of deep neural networks, it is difficult to extract easy to understand patterns from the model and discover causality relationship between the original dataset and outputs \cite{Ghorbani2018InterpretationON, Zhang2018OpeningTB}. The authors, however, have been exerting efforts to utilize statistical methods to extract such information from the trained models. These methods are usually based on gradient variation and are categorized as sample-based and model-based approaches, including Shap \cite{NIPS2017_7062, lundberg2018explainable}, DeepLift \cite{deeplift}, LIME \cite{lime},  and Interpret \cite{interpretML}. This can pave the way for finding novel biomarkers and regulatory interactions of different tissues and disease states.

\section{Acknowledgments}

Authors would like to acknowledge creative comments and ideas by Dr. S.M.Ali Eslami and Dr. Mahmoud Ghandi. We are also grateful to Hamid Malki for his artworks in Fig.~\ref{fig:architectures} and Fig.~\ref{fig:cls_hyperopt}.

\section{Author Contributions} 
ASZ and HC developed the idea of using DNNs for processing mRNA profiles. AS developed a pipeline for integration and preprocessing of the data. BA developed the deep learning framework and hyperparameter optimization, generated the results and created the figures. All authors contributed in discussing the results and writing the manuscript. HC provided the computational resources.

\section{Competing Financial Interests} The authors declare no competing   interests.
\bibliographystyle{naturemag}
\bibliography{papers,library}

\pagebreak
\section{Figure Captions}

\pagebreak

\renewcommand{\thetable}{S\arabic{table}}

\begin{table}
	\caption{The number of samples used for each type of tissue and cancer.}
	\label{tbl:numsamples}
	\begin{center}
		\includegraphics[width=\linewidth]{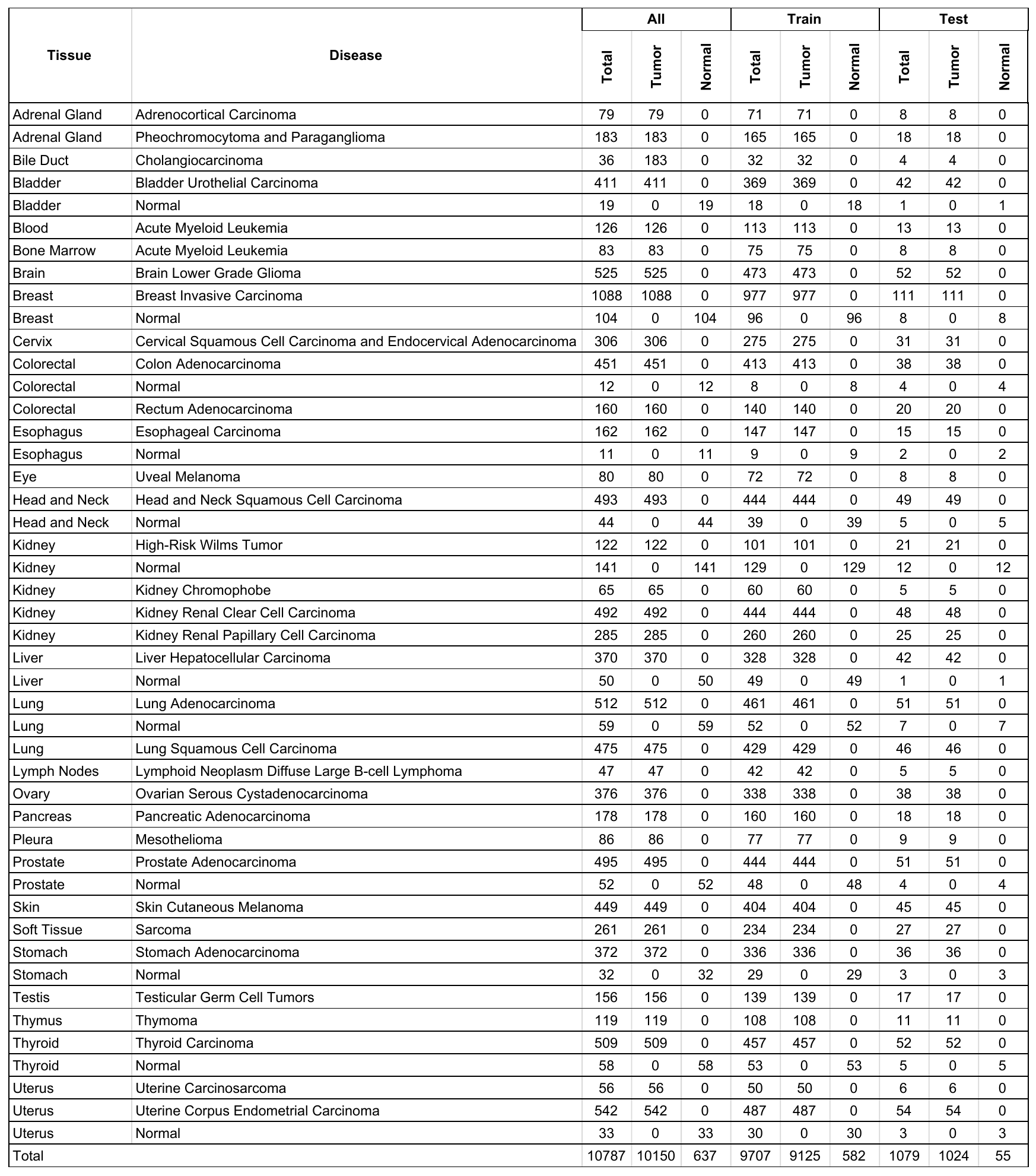}
	\end{center}
\end{table}

\begin{table}
	\caption{Sensitivity, specificity, F1 metric and balanced accuracy of DNN classification for each tissue.}
	\label{tbl:tissuef1}
	\begin{center}
		\includegraphics[]{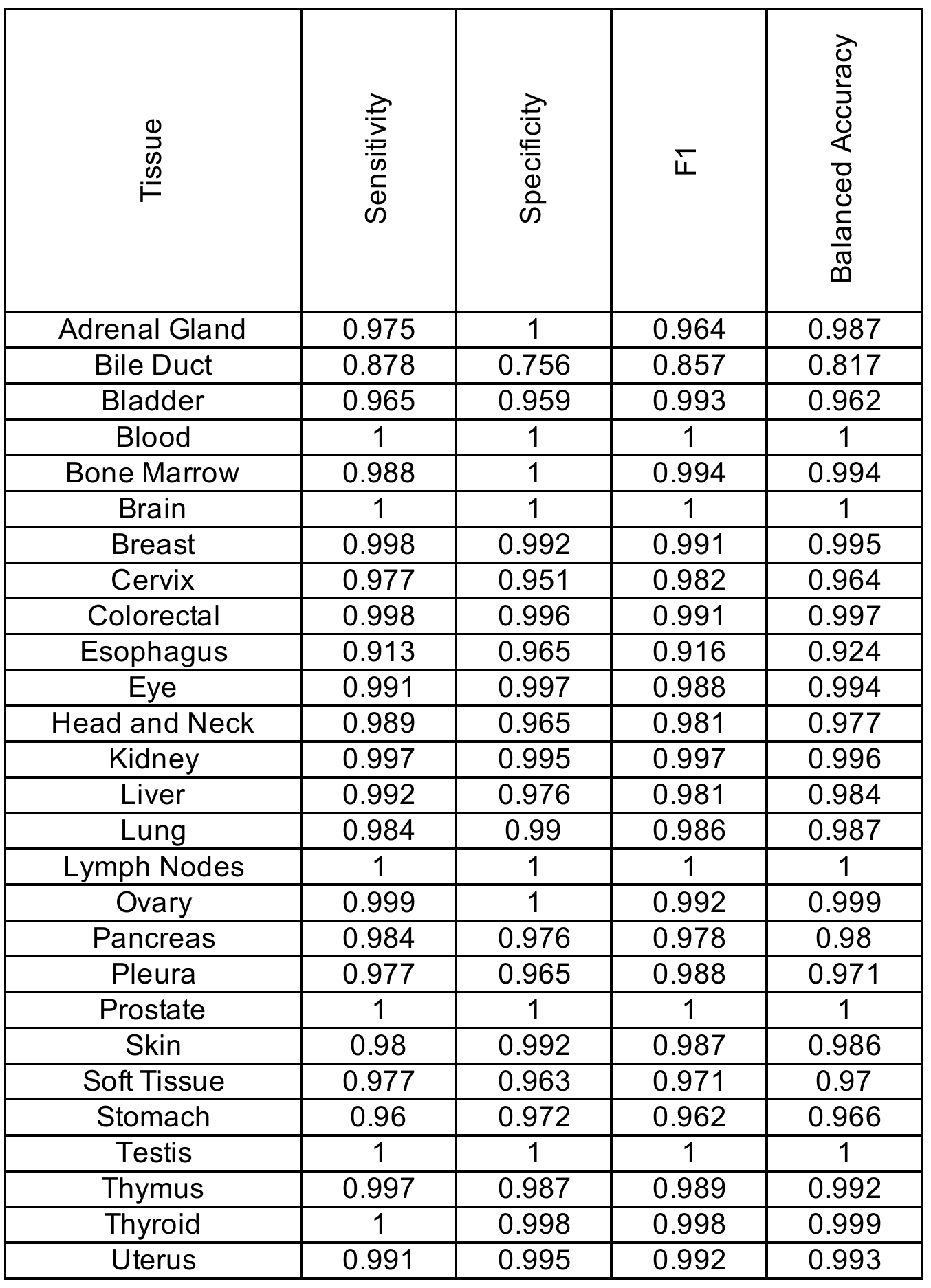}
	\end{center}
\end{table}

\begin{table}
	\caption{Sensitivity, specificity, F1 metric and balanced accuracy of DNN classification for each disease type.}
	\label{tbl:diseasef1}
	\begin{center}
		\includegraphics[]{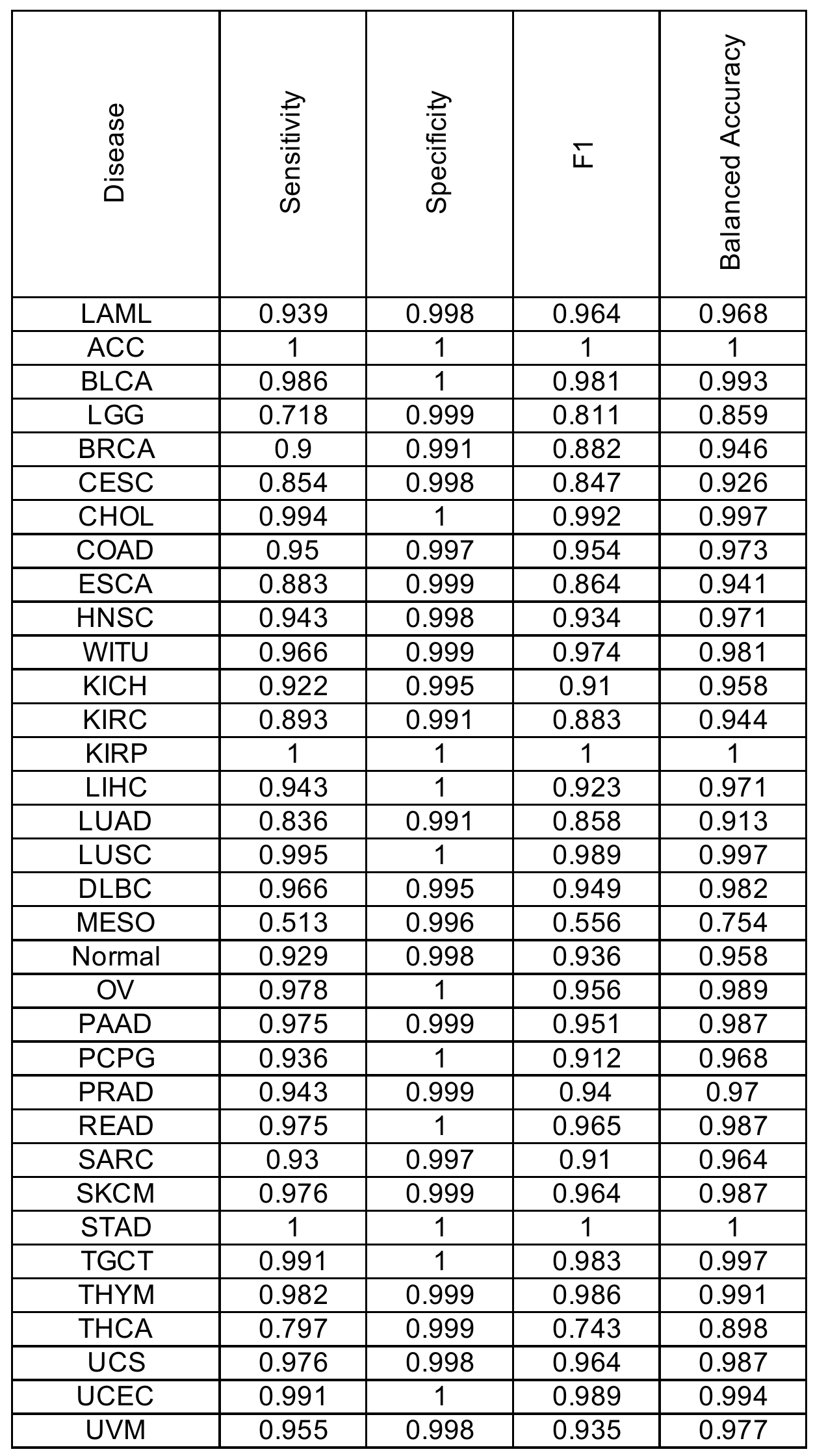}
	\end{center}
\end{table}

\renewcommand{\thefigure}{S\arabic{figure}}

\begin{figure}[h!]
	\begin{center}
		\includegraphics[width=.9\textwidth]{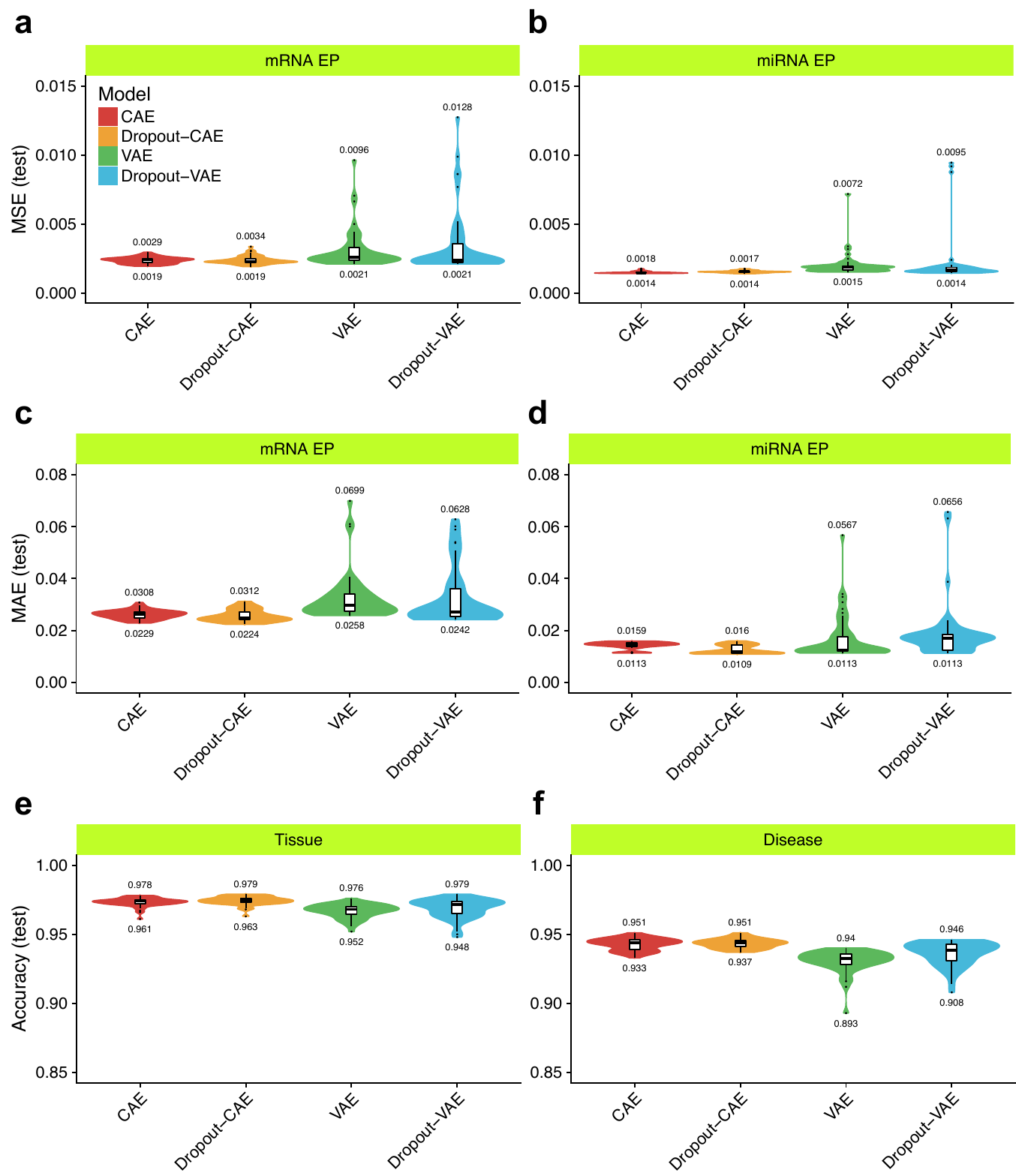}
	\end{center}
	\caption {Distribution of different error or accuracy measurements during hyperparameter optimization: \textbf{(a)} mean square error (MSE) of reproducing mRNA expression profiles (EP), \textbf{(b)} MSE of generating miRNA expression profiles, \textbf{(c)} mean absolute error (MAE) of reproducing mRNA EP, \textbf{(d)} MAE of reproducing miRNA EP, \textbf{(e)} accuracy of predicting tissue for the test dataset, \textbf{(f)} accuracy of predicting cancer type for the test dataset. In each violin plot, the colors represent different architectures.}
	\label{fig:hyperopt_dis} 
\end{figure}

\begin{figure}[h!]
	\begin{center}
		\includegraphics[width=0.7 \textwidth]{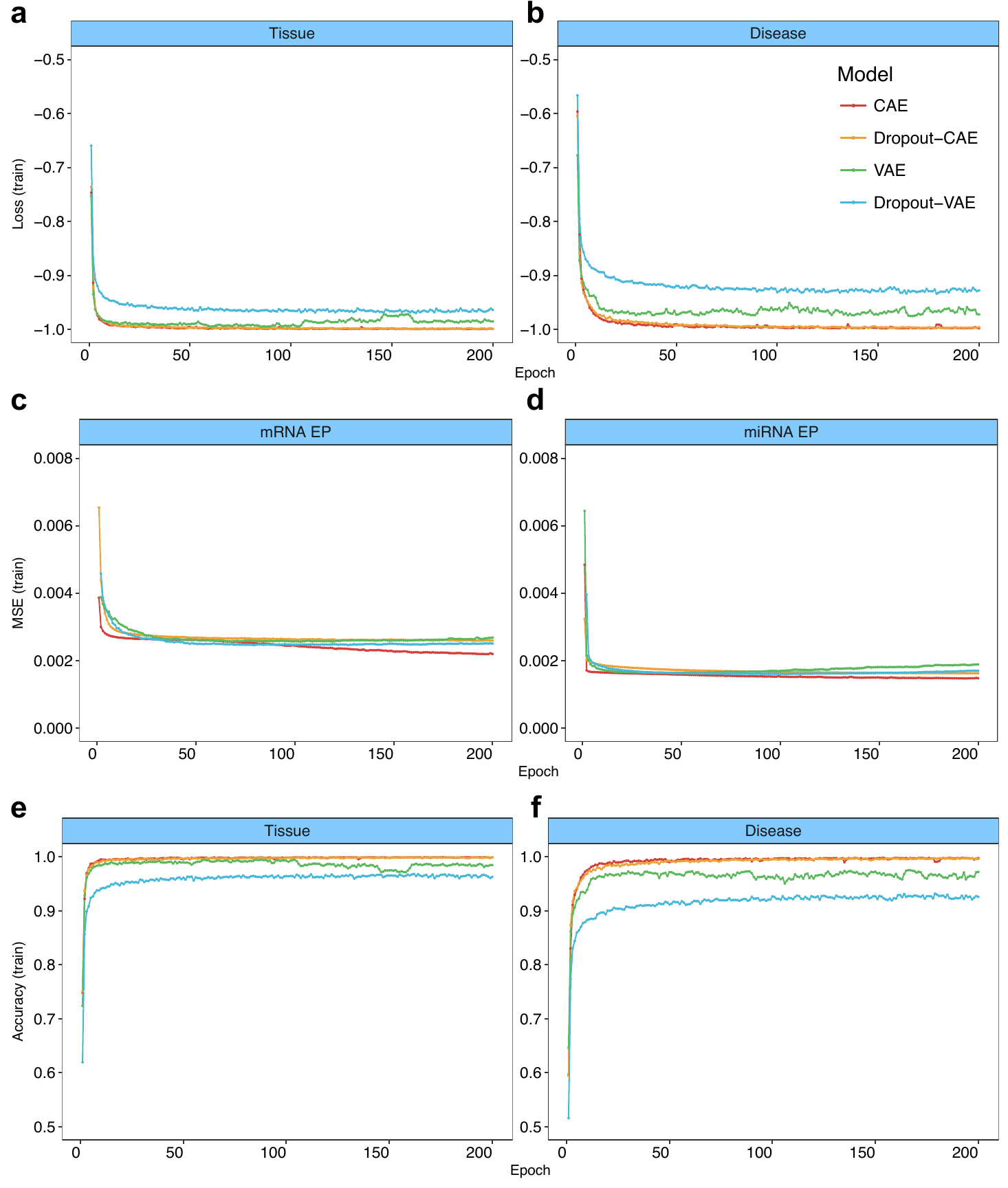}
	\end{center}
	\caption {Performance of DNNs on training data, during 200 epochs of training. In each plot, the x-axis shows the training epochs, and the y-axis shows: \textbf{(a)} the value of loss function for predicting tissue type, 
		\textbf{(b)} the value of loss function for predicting disease state,
		\textbf{(c)} mean square error (MSE) of reproducing mRNA expression profiles (EP), \textbf{(d)} MSE of predicting miRNA EP, \textbf{(e)} accuracy of predicting tissue, and \textbf{(f)} accuracy of predicting cancer type. All results are based on the training dataset.
	}
	\label{fig:hyperopt_train} 
\end{figure}

\begin{figure}[h!]
	\begin{center}
		\includegraphics[width=0.7 \textwidth]{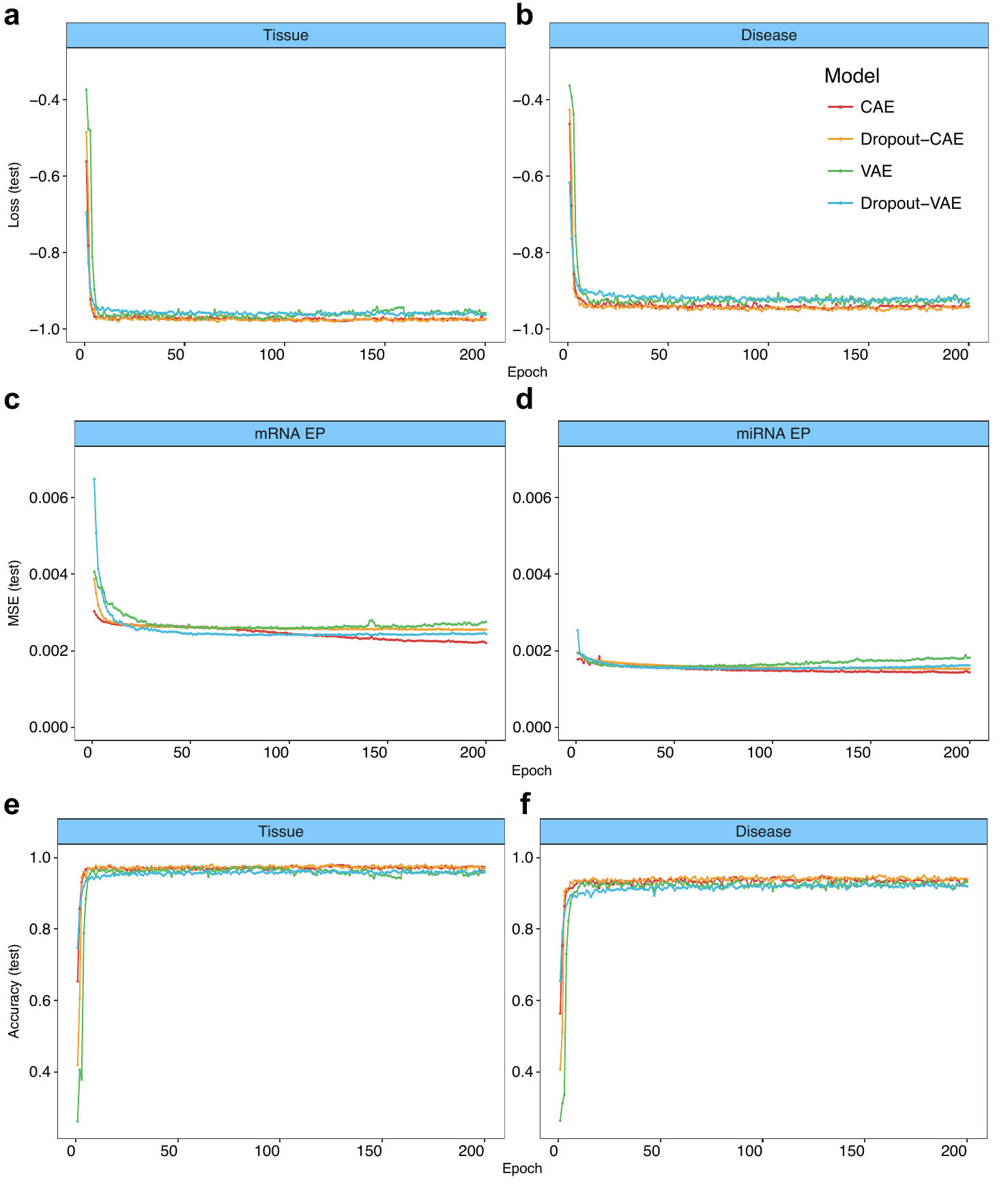}
	\end{center}
	\caption {Performance of DNNs on test data, during 200 epochs of training. In each plot, the x-axis shows the training epochs, and the y-axis shows: \textbf{(a)} the value of loss function for predicting tissue type, 
		\textbf{(b)} the value of loss function for predicting disease state,
		\textbf{(c)} mean square error (MSE) of reproducing mRNA expression profiles (EP), \textbf{(d)} MSE of predicting miRNA EP, \textbf{(e)} accuracy of predicting tissue, and \textbf{(f)} accuracy of predicting cancer type. All results are based on the test dataset.
	}
	\label{fig:hyperopt_test} 
\end{figure}

\begin{figure}[h!]
	\begin{center}
		\includegraphics[width=0.7\textwidth]{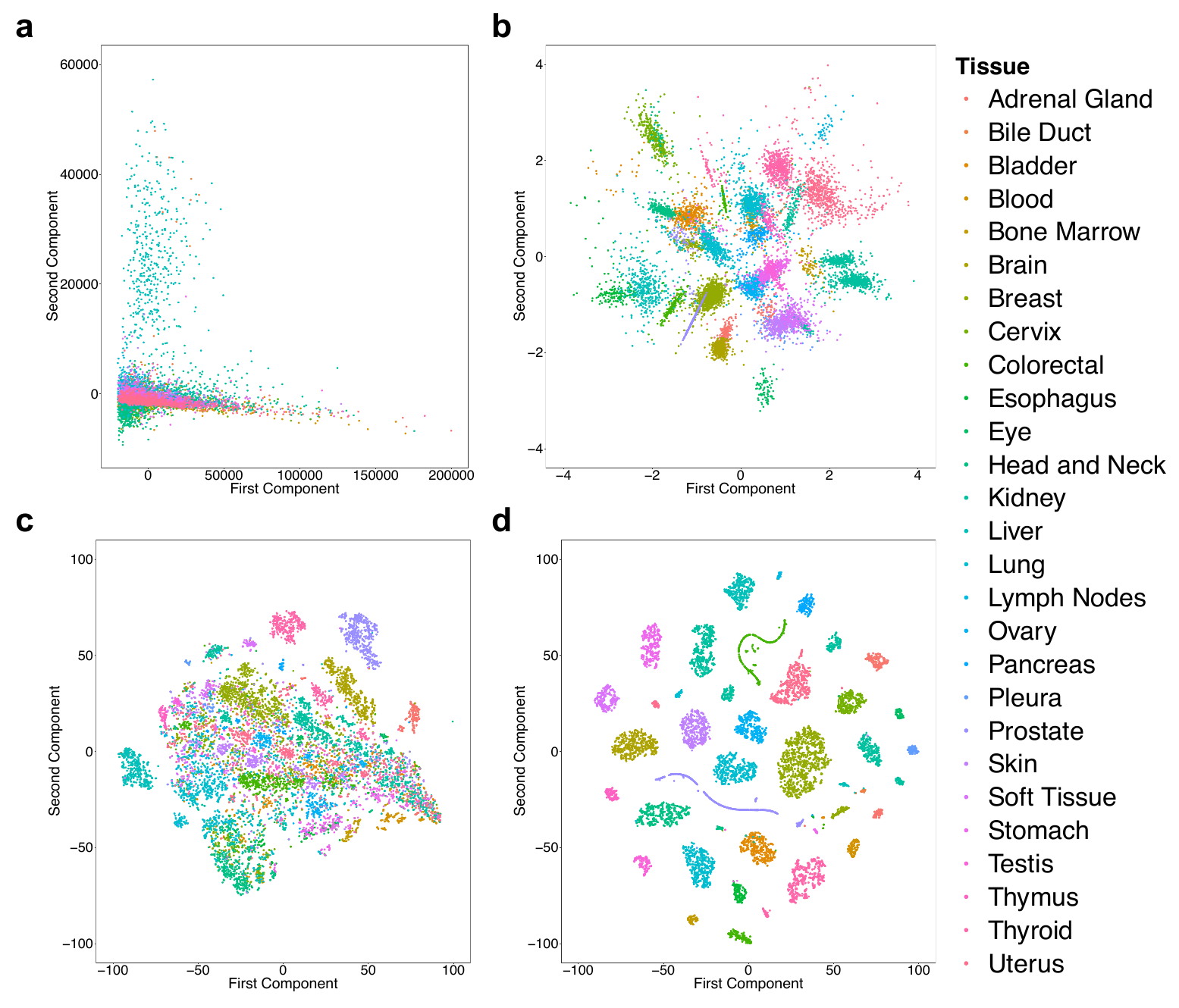}
	\end{center}
	\caption {Discrimination of sample tissues in the original and Cell Identity Codes (CIC) spaces \textbf{(a)} PCA plot of the original mRNA expression profiles of all samples. Each dot and its color show a sample and its tissue type, respectively. \textbf{(b)} PCA plot of the 8-dimensional CIC space.  \textbf{(c)} t-SNE plot of the original mRNA expression profiles.  \textbf{(d)} t-SNE plot of the 8-dimensional CIC space.
	}
	\label{fig:tsne_tissue} 
\end{figure}

\begin{figure}[h!]
	\begin{center}
		\includegraphics[width=0.7 \textwidth]{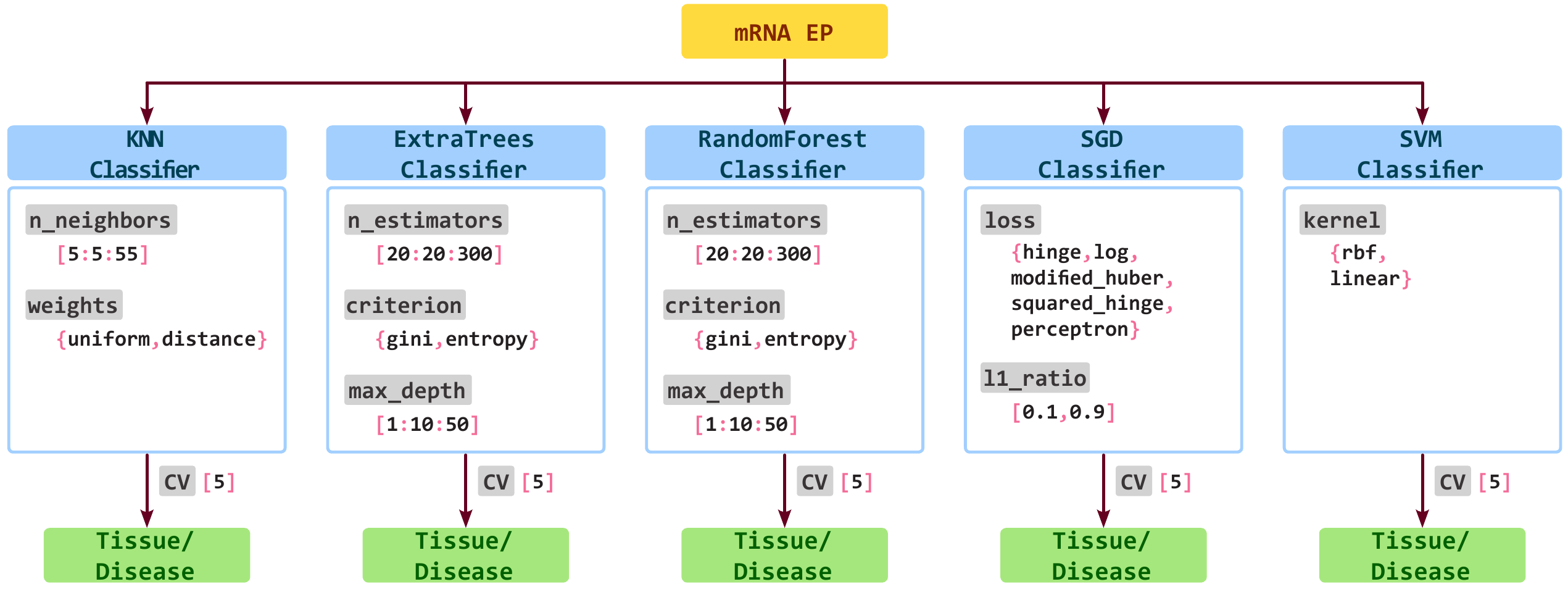}
	\end{center}
	\caption {Hyperparameter optmization of the other classification algorithms. Here the notation $[start:step:end]$ return evenly spaced values within the close interval $[start, stop]$ with increments equal to $step$. A dictionary $\{a, b, c\}$ means that all of the item $a$, $b$, and $c$ can be selected in the Bayesian optimization process.}
	\label{fig:cls_hyperopt} 
\end{figure}

\begin{figure}[h!]
	\begin{center}
		\includegraphics[width=0.7 \textwidth]{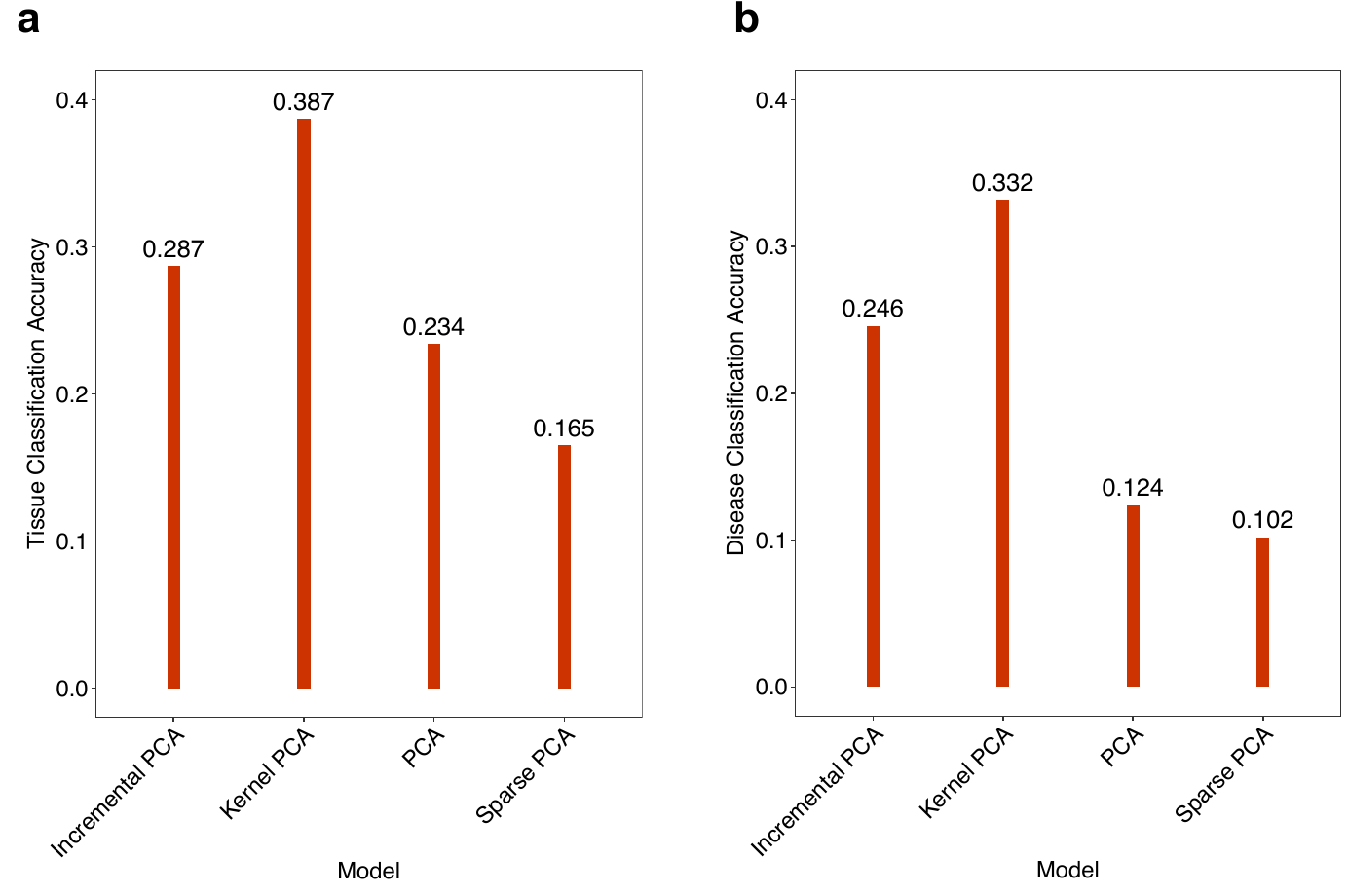}
	\end{center}
	\caption {	Accuracy of tissue 	\textbf{(a)} and disease \textbf{(b)} classification using traditional dimension reduction algorithms. }
	\label{fig:pca} 
\end{figure}
\end{document}